# Inspect Transfer Learning Architecture with Dilated Convolution

Syeda Noor Jaha Azim, Md. Aminur Rab Ratul
snjazim@uwaterloo.ca, mratu076@uottawa.ca

*Abstract*—**There are many award-winning pre-trained Convolutional Neural Network (CNN), which have a common phenomenon of increasing depth in convolutional layers. However, I inspect on VGG network, which is one of the famous model submitted to ILSVRC-2014, to show that slight modification in the basic architecture can enhance the accuracy result of the image classification task. In this paper, We present two improve architectures of pre-trained VGG-16 and VGG-19 networks that apply transfer learning when trained on a different dataset. I report a series of experimental result on various modification of the primary VGG networks and achieved significant out-performance on image classification task by: (1) freezing the first two blocks of the convolutional layers to prevent over-fitting and (2) applying different combination of dilation rate in the last three blocks of convolutional layer to reduce image resolution for feature extraction. Both the proposed architecture achieves a competitive result on CIFAR-10 and CIFAR-100 dataset.**

## I. INTRODUCTION

Convolutional networks (ConvNets) have achieved excellent success in the large-scale image and video recognition, which has become feasible before large public image repositories such as ImageNet [1] and high-performance computing systems such as GPUs or large-scale distributed clusters. These advancements were largely motivated by strong baseline schemas, such as semantic segmentation [2], object recognition [3], image captioning [4], and human pose estimation[4]. What all the latest Convolutional Neural Networks(CNN) architectures have in common is the growing depth and complexity in the networks that offer better accuracy. While this strategy has been beneficial, some inevitable problems occur when the network becomes more complicated. Cost and overhead computing and memory utilization is one of the critical problems generated by excessive effort to make networks deeper and more complicated to make them perform better.

Because of the cost of data acquisition and expensive an-notation, which limits its development, it is tough to build a well-annotated large-scale dataset. Transfer learning [5] re-laxes the hypothesis that the training data must be independent and identically distributed with the test data, which motivates to use of transfer learning to address the issue of inadequate training data. In the transfer learning, target domain model need not be trained from scratch, which can significantly reduce the demand for training data and training time in the target domain. The main contribution of my work is to propose a simple architecture that applies network-based deep transfer learning with predictive modeling problems using fine-tuned VGG [6] architectures. VGG architecture is built on ImageNet [1] by Karen Simonyan and Andrew Zisserman for the purpose of image recognition and classification. ImageNet is a dataset that contains 14 million images with over 1,000 classes. In this project, image classification is experimented by adding multiple dilation rates in a different layer of the VGG network, which results in better accuracy. To evaluate this technique, I conduct tests on popular image recognition datasets CIFAR-10 and CIFAR-100.

This paper is organized as follows. Section II discusses the related work on transfer learning and dilated convolutional models. Section III reviews the methods used for feature extraction and the presented architectures. Section IV is an elaboration of software used, data structures, program structures, data representation and any special setup needed. Finally, Section V reports a series of experimental results on CIFAR datasets for varying dilation rates in convolutional layers. Finally, conclusions and future work are summarized in Section VI, and acknowledgment is covered at the end of this paper.

## II. LITERATURE REVIEW

Deep convolutional networks have a strong history of computer vision, demonstrating excellent outcomes in using supervised back-propagation systems to conduct digit recognition [7]. With the more recent advancement, these networks have accomplished competitive results on significant benchmark datasets composed of more than one million images, such as ImageNet [1], especially the convolutional network suggested by Krizhevsky et al. (2012) [3].

There is also a brief history of learning from associated tasks in machine learning, starting with [8] and [9]. Later works like [10] established effective frameworks to optimize models from related tasks, and [11] researched how parameter manifolds can be transferred to current tasks [12]. A main issue for such learning challenges is to discover a representation of features that captures details specific to the object category while discarding noise irrelevant to data about object categories such as brightness. It is feasible to improve the feature extraction stage of deep learning methods by 'transfer learning', which consists of tuning the parameters trained in one feature space to operate in another feature space. Some methods that use transfer learning are: Spatial pooling [13], MOP-CNN [14], Neural codes [15] and R-MAC [16]. CNN extract data-driven characteristics from input data (e.g. image, video, and audio information) that is structured in regular typically low-dimensional grids (See Fig.1[17]). In order to facilitate the modeling process, such grid structures are often assumed to have statistical features (e.g., stationary and locality). Learning algorithms then use this hypothesis and increase efficiency by decreasing parameter complexity.

Models based on dilated convolution have been actively studied for semantic segmentation but very few for image classification along with transfer learning. For instance, [18] experiments with the impact of changing dilation rates to capture long-range data, [19] adopt hybrid dilation rates within ResNet's last two blocks, while [20] further suggests learning the deformable convolution that samples input characteristics with learned offset, generalizing dilated convolution. [21] uses image captions, [22] uses video gestures, and [23] includes depth information to enhance the precision of the segmentation model further. Besides, [24, 25, 26] has applied dilated convolution to object detection but have not applied transfer learning.

## III. METHODS OVERVIEW

In this section, I describe how dilated convolution is implemented in fine-tuned VGG networks to extract dense features for image classification and recognition. And then discuss the proposed architecture with dilated convolution utilized in parallel.

### A. Dilated Convolution for Dense Feature Extraction

Dilated convolution, also known as Atrous convolution, was initially designed for the efficient computation of the in-comparable wavelength transform in the "algorithme a` trous" scheme of [27]. To reduce feature resolution, I would prefer to use dilated convolution instead of Deep Convolutional Neural Networks (DCNNs) [7].

In a two-dimensional signal the dilated convolution is applied over the input feature map x for each location i on the output y and a filter w:

$$y[i] = \sum_k x[i + r.k]w[k] \quad (1)$$

Eq.1 [28] denotes input signal is sampled with the dilation(atrous) rate r that corresponds to the stride, which is equivalent to convolving the input x with upsampled filters produced by inserting (r -1) zeros (holes) between two consecutive filter weights along each spatial dimension (hence the name atrous convolution where the French word trous means holes in English). Dilated convolution enables the field-of-view filter to be adapted by altering the rate value, where else standard convolution is a special case for rate r = 1. See Fig.2 [19] for illustration.

### B. Convolutional Layer

Detecting local conjunctions of features from the previous layer and mapping their appearance to a feature map [29] is the main task of the convolutional layer. The image is fragmented into perceptrons as a result of convolution in neural networks, creating local receptive fields and ultimately compressing the perceptrons in $m_2$ $m_3$ feature maps. This map, therefore, stores the data where the distinguishing feature appears in the image and how well it matches the filter. Each filter is therefore learned in spatial in terms of the position in the volume to which it is applied. There is a bank of $m_1$ filters in each layer. The amount of how many filters in one stage are applied is equal to the volume depth of the feature maps of the output. Each filter detects a specific feature at each input location. The output $y_i^{(l)}$ of layer l consists of $m_i^{(l)}$ feature maps of size $m_2^{(l)} \times m_3^{(l)}$. The i$^{th}$ feature map, known as $y_i^{(l)}$, is calculated as eq.2 [30]. Where $B_i^{(l)}$ is a bias matrix and $k_{i,j}^{(l)}$ is the filter.

$$y_i^{(l)} = B_i^{(l)} + \sum_{j=1}^{m_1^{(l-1)}} k_{i,j}^{(l)} * y_j^{(l-1)} \quad (2)$$

### C. Pooling Layer

It is the responsibility of the pooling or down-sampling layer to reduce the spatial size of activation maps. They are generally used after various phases of other layers (i.e., convolutional and non-linearity layers) to gradually decrease computing demands through the network and minimize the probability of over-fitting. The pooling layer has two hyperparameters, the spatial extent of the filter $F^{(l)}$ and the stride $S^l$. It takes an input volume of size $m_1^{(l-1)} \times m_2^{(l-1)} \times m_3^{(l-1)}$ and provides an output volume of size $m_1^{(l)} \times m_2^{(l)} \times m_3^{(l)}$ as follows [31]:

$$m_1^{(l)} = m_1^{(l-1)}$$

$$m_2^{(l)} = (m_2^{(l-1)} - F^l)/S^l + 1$$

$$m_3^{(l)} = (m_3^{(l-1)} - F^l)/S^l + 1$$

The main idea of the pooling layer is to provide translational invariance as feature detection is more essential compared to the accurate location of the feature, especially in image identification tasks. The pooling procedure, therefore, seeks to maintain the identified characteristics in a narrower representation by discarding less important information at the expense of spatial resolution.

### D. Deep architecture for image classification

I adapt the two ImageNet pre-trained [1] VGG [6] architectures for image classification by applying dilated convolution to extract dense features. Different dilation rate is added to the blocks of VGG architecture, and $32 \times 32 \times 3$ RGB image is used as input to train the proposed architecture. Motivated by multi-grid methods utilizing a hierarchy of grids of distinct dimensions [32, 33, 34, 35], I implement varying dilation rates in the presented architecture within block 3 to block 5.

1) **Dilated Convolution in VGG-16 Architecture**: For VGG-16, the first two blocks(each with 2 convolutional layers) of the network are frozen to ensure the model to be computationally inexpensive. Since the model is pre-trained over ImageNet, so the model already learned the feature and colour combinations of many input images. Therefore, training the entire model leaves the chance of over-fitting, i.e., training images are learned so well that the model fails to detect new images from the test set.

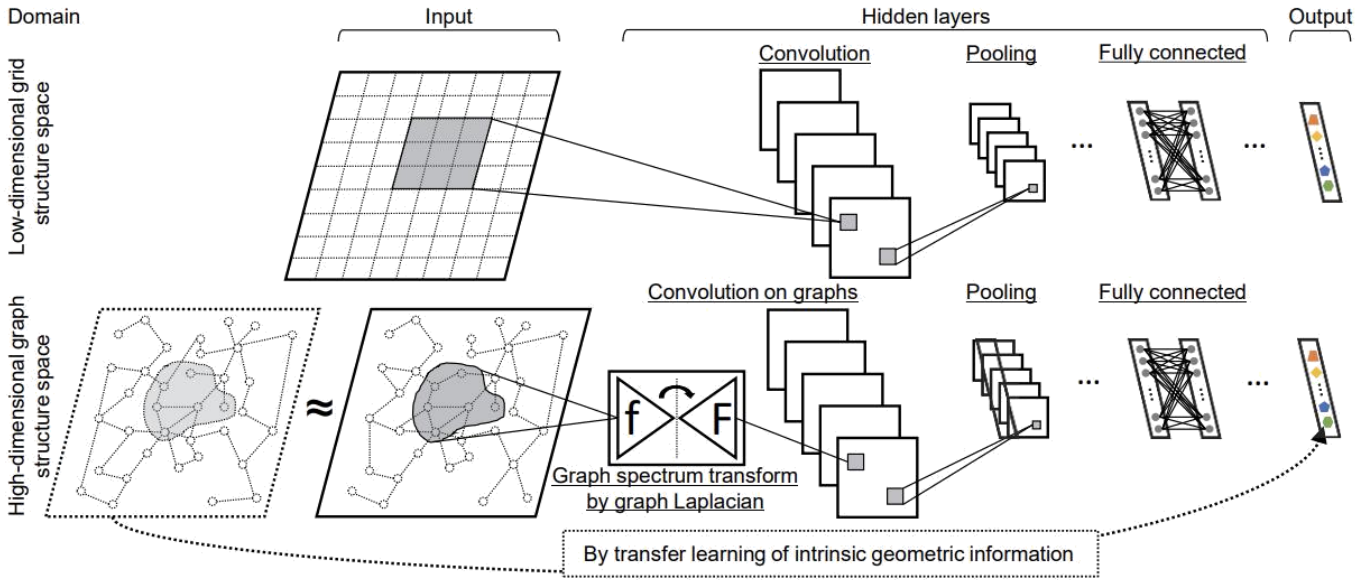

*Fig. 1. Conventional CNN operates on a periodic grid domain (top); suggested CNN learning structure that can transfer inherent geometric information from a source graph domain to a target graph domain (bottom).[17]*

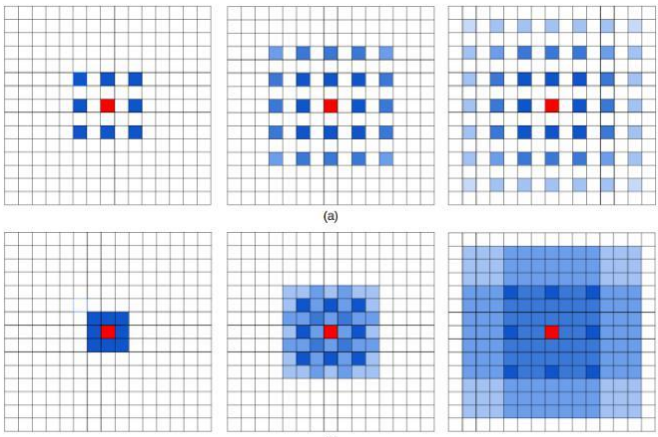

*Fig. 2. Left to right: the pixels (marked in blue) contribute to the center pixel calculation (marked in red) through three kernel size 3 convolution layers. (a) dilation rate r = 2 for all convolutional layers. (b) the dilation rates of subsequent convolution layers are r = 1, 2, 3, respectively. [19]*

The third block of the VGG-16 network uses three 3 3 convolutions (Conv.) layers with filter size 256, followed by a max-pooling layer for down-sampling. I initiated dilation rate 2 on the third block to achieve better accuracy where the filter size is 256. Dilation rate 4 is adopted for the fourth block of the network, which also has three 3 3 Conv layers but with 512 filter size. In the last block of the VGG-16 convolutional layer, two dilation rates 4 and 8 are concatenated simultaneously.

The extracted feature output is flattened and feeds the resulting image vector to the first 1 1 fully-connected layer with a 512 filter size for image classification. Another 1 1 fully-connected layer with 256 filter size is aggregated to reduce the image size gradually. Each of these fully-connected layers is equipped with the rectification (ReLU) non-linearity function
[36]. Finally, the last 1 1 fully-connected layer with 10 (for CIFAR-10) or 100 (for CIFAR-100) filter size is used with soft-max function to achieve the classified image. The dilated proposed architecture for VGG-16 network is illustrated in Fig.3

  2) **Dilated Convolution in VGG-19 Architecture**: VGG-19 is a 19 layers deep convolutional neural network. Changes in architecture that I have made in the VGG-19 is the same as the change in the architecture of VGG-16, apart from the different dilation rate applied for the block. I have also frozen the first 2 blocks of the VGG-19 model so that the proposed model does not update weight from the layers of these two blocks via back-propagation and cause over-fitting. Then, initiated dilation rate 2 on the third block which has four 3 3 Conv layers with filter size 256. The same dilation rate is adopted for the fourth block of the network, which also has four 3 3 Conv. layers but with a 512 filter size. For the last convolutional block, which is the same as the fourth block, dilation rate 2 and 4 is concatenated side by side to maximize the performance of feature extraction. Each block of the convolution layer is followed by a max-pooling layer. Hereafter, the flattened image is propagated to three 1 1 fully-connected layers with different filter size. The first fully-connected layer has a filter size 512 and the second fully-connected layer has a filter size of 256, but both are ReLU activation function. The last fully-connected layer has filter size of 10 or 100 depending on which dataset I am training the model. Finally, the classified image is obtained from the last fully-connected layer, which is equipped with soft-max function. The dilated proposed architecture for the VGG-19 network is illustrated in Fig.4.

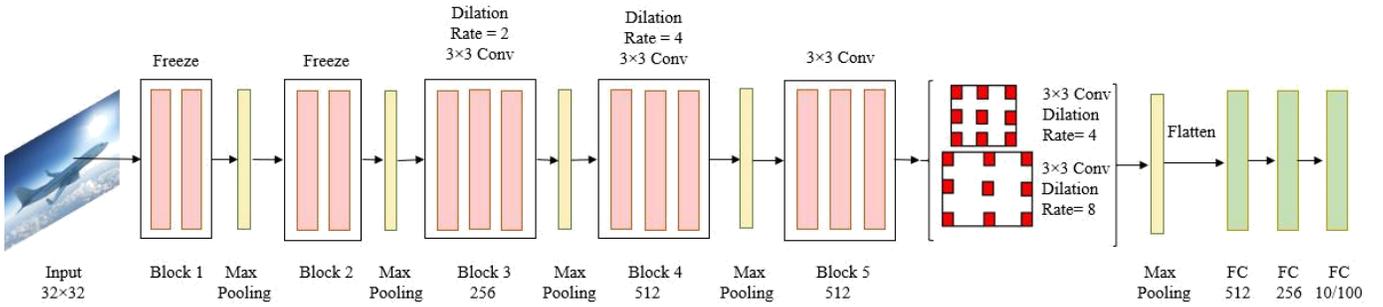

*Fig. 3. Dilated architecture of VGG-16 network freeze Block 1 and Block 2. Apply dilation rate 2 in Block 3 and 4 in Block 4. Concat dilation rate 4 and 8 in Block 5. Flatten the extracted image and propagate it to 3 fully-connected(FC) layer with filter size 512, 256, and finally 10 or 100 depending on the dataset.*

## IV. IMPLEMENTATIONS

My implementation is based on two model VGG-16 and VGG-19. For a fair comparison, I have trained both base architectures and proposed an architecture for CIFAR-10 [37] and CIFAR-100 [37] dataset. Each dataset is split into 40,000 training sets, 10,000 validation set, and 10,000 test set. To train each model, 250 iterations were performed with a learning rate of $10^{-5}$. Validation loss is monitored for every 7 iterations and if there is no reduction in the validation rate the learning rate has been decreased by sqrt(0.05). This process is repeated for every 7 epoch for effective training of the proposed architecture. Further escalation of learning is obtained by applying a straight-forward Adam [38] optimizer, an algorithm for first-order gradient-based optimization of stochastic objective functions. Categorical Cross-Entropy [39] loss function (negative log-likelihood) is implemented to achieve a probabilistic interpretation of the accurate result.

In order to acquire the result of the experiment, Keras [40] is used to implement the front-end of the program, which is an open-source neural network library written in Python. For the back-end Tensorflow [41], which is developed by Google Brain team, the Library is used. The proposed model run in Core i7 - 8750H processor with 2.2 GHz and NVIDIA GeForce GTX 1050Ti GPU.

## V. EXPERIMENTAL RESULTS

### A. Dataset and evaluation methodology

I use two public balanced-datasets CIFAR-10 and CIFAR-100 which are 80 million tiny labeled image dataset subsets gathered by Alex Krizhevsky, Vinod Nair, and Geoffrey Hin-ton.

CIFAR-10 [37] dataset consists of 60,000 32 32 color images in 10 classes, with 6000 images per class. There are 50,000 images of training and 10,000 images of testing. The dataset is split into five training batches and one test batch with 10,000 images each. The test batch includes precisely 1000 randomly selected images from each class. The training batches contain the remaining images in random order, but some training batches may contain more images from one class than another. The training batches contain precisely 5000 images of each class. Fig.5 shows 10 sample images from each class of CIFAR-10 dataset.

CIFAR-100 [37] dataset is the same as the CIFAR-10, except that there are 100 classes with 600 images each. There are 500 images of training and 100 images of testing per class. The 100 classes of CIFAR-100 are divided into 20 super-classes. Each image comes with a "fine" label (the class it belongs to) and a "coarse" label (the super-class it belongs to). Fig.6 shows sample images from the CIFAR-100 dataset.

Data Prepossessing. Inspired by AutoAugmentation [42], which is a new automated data augmentation technique presented by Google, I have applied data augmentation to significantly increase the diversity of data available for training models. The pre-processing baseline follows the convention: standardization of data, use of vertical and horizontal random flips, rotation range 30, width and height shift range l0.3 and zoom range 0.3. Finally, 40,000 training images results in $10^7$ new images after running 250 epochs, which enables the model to learn on more varieties of images.

Evaluation Methods. I followed the standard accuracy metric for the image classification evaluation procedure. Accuracy means how many data points are predicted correctly. After feeding different combinations of dilation rate in the convolutional layers of VGG-16 and VGG-19 networks I have trained each setup of the architecture with two datasets CIFAR-10 and CIFAR-100 to compare the accuracy of image classification.

### B. Result and Discussion

Table. I illustrate the performance of the different combinations of dilation rates applied to two VGG networks. VGG is a pre-trained fine-tuned model where every layer has the weight of the ImageNet database. When I train the model for CIFAR-10 or CIFAR-100 dataset it learns the weight of these datasets and adds new weight to each convolutional layer, this is how transfer learning model like VGGNet transfers the learned data from one dataset and utilizes it to train the model for the new dataset. This is how a pre-trained deep learning network out-perform other generic deep learning networks.

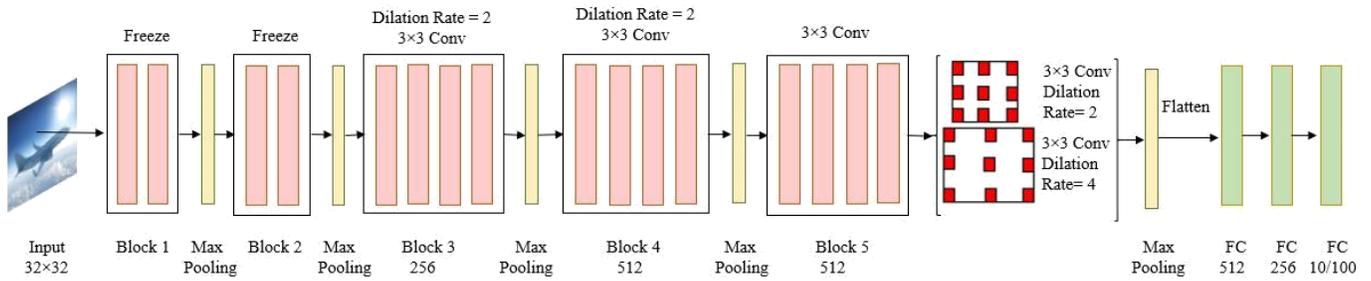

*Fig. 4. The dilated architecture of VGG-19 network freeze Block 1 and Block 2. Apply dilation rate 2 in Block 3 and 4. Concat dilation rate 2 and 4 in Block 5. Flatten the extracted image and propagate it to 3 fully-connected(FC) layer with filter size 512, 256, and finally 10 or 100 depending on the dataset.*

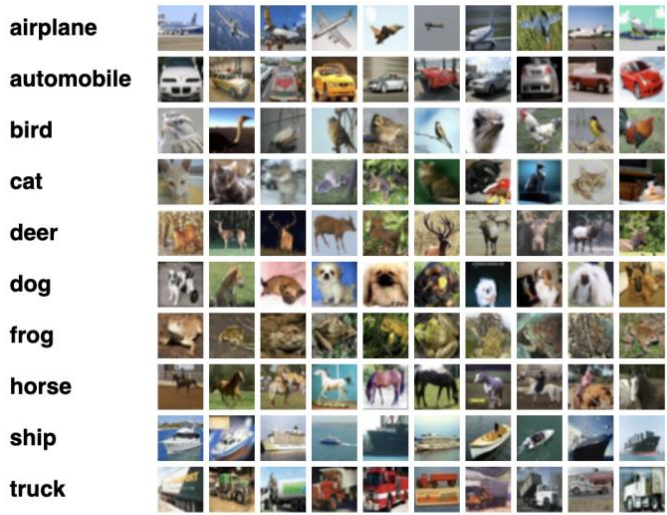

*Fig. 5. CIFAR-10 dataset with 10 random images from each class. [37]*

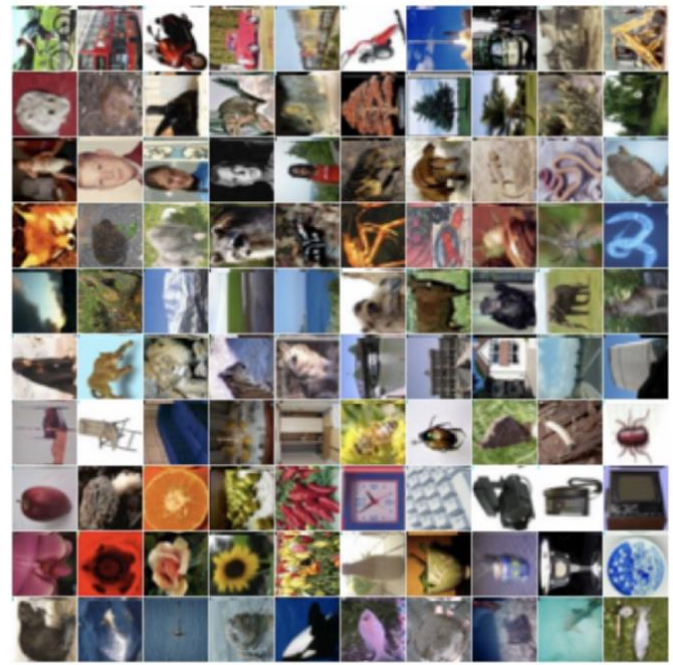

*Fig. 6. CIFAR-100 dataset with random images from each class. [37]*

VGG-16 trained on CIFAR-10. First, I have trained the basic model of VGG-16 for Cifar-10 dataset and compared the accuracy(%) result with the other modified VGG-16 networks. Freezing the layers (means not changing the weights during gradient descent or optimization) of the first block and adding dilation rate to the rest of the blocks in a hierarchical manner by a factor of 2 as shown in Table. I result in a reduction in accuracy percentage. Later, I freeze the first two blocks of the dilated VGG-16 network, which results in an increase in accuracy results by 2.8%. Finally, I concat dilation rates 4 and 8 and applied to each convolutional layer of the fifth and the last block of the network. The addition of a concatenated dilation rate in VGG-16 architecture out-performs the basic VGG-16 model by 6%.

VGG-16 trained on CIFAR-100. The proposed modified VGG-16 model is also trained for CIFAR-100 dataset to ensure that the proposed architecture out-performs for different dataset with more categories. For the CIFAR-100 dataset, the presented model out-performs the basic VGG-16 model by 4.3% accuracy. Therefore, the proposed architecture will perform better than the basic VGG-16 network for any dataset.

VGG-19 trained on CIFAR-10 and CIFAR-100. To validate the concept of adopting dilation in the convolutional layer, I have experimented with the VGG-19 network which has deeper convolutional layers. Similarly, I have compared the performance of the proposed model trained on CIFAR-10 with the basic VGG-19 system. Different combination of dilation rate is applied in VGG-19 network to achieve better performance on the image classification task. The gradual change in accuracy result is observed for various combinations of dilation rates in each convolutional layer of the VGG-19 network. As shown in the Table. I the best performance of VGG-19 network is observed when I freeze the convolutional layers of the first two blocks, add dilation rate 2 in the convolutional layers of block 3 and 4, and finally, apply dilation rate 2 and 4 in the convolutional layers of the last block. For CIFAR-10 dataset the proposed VGG-19 model shows 3.9% improvement in accuracy. When trained on the CIFAR-100 dataset, the proposed architecture outperforms the basic VGG-19 model by 6.5%.

Evaluation with other architecture. Comparative analysis of accuracy results for different architectures trained on CIFAR-10 is shown in Table.II. The accuracy percentage on the table shows the proposed architecture of VGG-16 outperforms proposed VGG-19 by 3.2% and Springenberg et al.[43] outperforms proposed VGG-16, approximately 2%. Therefore, it is clear that adding the dilation rate to the VGG networks enhances the performance of image classification when trained on the CIFAR-10 dataset.

State-of-art GPipe [51] outperforms proposed VGG-16 architecture trained on CIFAR-100 by 3.4%, as shown in Table.III. When VGG networks are trained on CIFAR-100 dataset proposed VGG-16 network outperforms the proposed VGG-19 network even though VGG-19 is a deeper convolutional network than VGG-16. Therefore, it can be concluded that increasing the convolutional layer in a network is not the only way to enhance the network's performance in the image classification task.

## VI. CONCLUSION

In this work, I applied dilation on convolutional layers of VGG-16 and VGG-19 networks to perform an image classification task. The basic pre-trained VGG networks learn image features from the ImageNet dataset and transfer learning of the extracted feature while training the model on the new dataset, like CIFAR-10/100. Freezing the first two blocks of each network prevented the over-fitting of the models. Adding a different combination of dilation rate to convolutional layers of the last three blocks shows better performance compared to the basic architecture. Nevertheless, the proposed architectures proved to be an excellent competitor to other more advanced techniques. For different recognition tasks and diverse datasets, the same trend was observed which highlights the effectiveness and generality of the learned representations.

In future work, I would like to apply a varied combination of dilation rate on other advanced pre-trained deeply con-voluted networks like GoogLeNet [51], Mask R-CNN [52] and YOLO9000 [53]. I would also experiment with the proposed architecture on several other public datasets like VisualQA [54], SVHN [55], IMDB Reviews [56], and WordNet [57] to validate out-performance of the proposed architecture on the image classification task.

TABLE I
EFFECT OF DIFFERENT COMBINATION OF DILATION RATE FOR VGG-16 AND VGG-19 ARCHITECTURE

| Network | Dilation Rate | | | | | CIFAR-10 | CIFAR-100 |
|---|---|---|---|---|---|---|---|
| | Block 1 | Block 2 | Block 3 | Block 4 | Block 5 | Accuracy(%) | Accuracy(%) |
| VGG-16 | Basic | | | | | 88.47 | 84.62 |
| VGG-16 | freeze | 1 | 2 | 4 | 8 | 86.19 | 83.5 |
| VGG-16 | freeze | freeze | 2 | 4 | 8 | 90.92 | 86.88 |
| VGG-16 (proposed) | freeze | freeze | 2 | 4 | 4, 8 | 93.75 | 88.28 |
| VGG-19 | Basic | | | | | 87.4 | 81.09 |
| VGG-19 | freeze | 1 | 2 | 4 | 8 | 87.95 | 83.79 |
| VGG-19 | freeze | freeze | 2 | 4 | 8 | 88.56 | 85.14 |
| VGG-19 (proposed) | freeze | freeze | 2 | 2 | 2, 4 | 90.85 | 86.38 |

TABLE II
COMPARISON OF ACCURACY RESULT OF CONVOLUTIONAL NETWORKS TRAINED ON CIFAR-10.

| Models | CIFAR-10 Accuracy(%) |
|---|---|
| Springenberg et al. [43] | 95.59 |
| VGG-16 (proposed model) | 93.75 |
| VGG-19 (proposed model) | 90.85 |
| Stollenga et al. [44] | 90.78 |
| APAC [45] | 89.67 |
| ReNet [46] | 87.65 |

TABLE III
COMPARISON OF ACCURACY RESULT OF CONVOLUTIONAL NETWORKS TRAINED ON CIFAR-100.

| Models | CIFAR-10 Accuracy(%) |
|---|---|
| GPipe [47] | 91.3 |
| VGG-16 (proposed model) | 88.28 |
| VGG-19 (proposed model) | 86.38 |
| DeVries et al. [48] | 84.4 |
| Liang et al. [49] | 83.8 |
| Huang et al. [50] | 82.82 |